\documentclass{article}

\usepackage[preprint, nonatbib]{nips_2018}

\usepackage{float}

\usepackage[square,numbers]{natbib}
\usepackage{makecell}
\usepackage[utf8]{inputenc} % allow utf-8 input
\usepackage[T1]{fontenc}    % use 8-bit T1 fonts
\usepackage{hyperref}       % hyperlinks
\usepackage{url}            % simple URL typesetting
\usepackage{booktabs}       % professional-quality tables
\usepackage{amsfonts}       % blackboard math symbols
\usepackage{nicefrac}       % compact symbols for 1/2, etc.
\usepackage{microtype}      % microtypography
\usepackage{graphicx}
\usepackage{tabularx}
\usepackage{hyperref}
\usepackage{multirow}

\title{QACP: An Annotated Question Answering Dataset for Assisting Chinese Python Programming Learners}

\author{
  Rui Xiao \\
  \texttt{xiaorui@pku.edu.cn}\\
\And
 Lu Han \\
  \texttt{lu.han@bdqn.cn} \\
\And
 Xiaoying Zhou \\
  \texttt{xiaoying.zhou@bdqn.cn} \\
\AND
 Jiong Wang \\
  \texttt{2211002040@cnu.edu.cn} \\
\And
 Na Zong \\
  \texttt{na.zong@bdqn.cn} \\
\And
 Pengyu Zhang \\
  \texttt{pengyu.zhang@bdqn.cn} \\
}

\begin{document}
\maketitle

\begin{abstract}

In online learning platforms, particularly in rapidly growing computer programming courses, addressing the thousands of students' learning queries requires considerable human cost. The creation of intelligent assistant large language models (LLMs) tailored for programming education necessitates distinct data support. However, in real application scenarios, the data resources for training such LLMs are relatively scarce. Therefore, to address the data scarcity in intelligent educational systems for programming, this paper proposes a new Chinese question-and-answer dataset for Python learners. To ensure the authenticity and reliability of the sources of the questions, we collected questions from actual student questions and categorized them according to various dimensions such as the type of questions and the type of learners. This annotation principle is designed to enhance the effectiveness and quality of online programming education, providing a solid data foundation for developing the programming teaching assists (TA). Furthermore, we conducted comprehensive evaluations of various LLMs proficient in processing and generating Chinese content, highlighting the potential limitations of general LLMs as intelligent teaching assistants in computer programming courses.\footnote{The dataset can be found at \url{https://github.com/NTAIX/Chinese-Python-QA-Dataset}} 

\end{abstract}

\section{Introduction}
With the impressive performance of generative artificial intelligence (Generative AI) applications, such as ChatGPT \cite{openai2023chatgpt}, there is widespread attention and deep interest in the potential and application of large language models (LLMs). In the field of education, artificial intelligence (AI) is rapidly transforming the way computer science (CS) is taught and learned. In this background, the development of generative AI offers opportunities for AI to play the role on assisting to teach programming courses \cite{lee2023learning}.

For example, well-known AI assistants such as ChatGPT and GitHub Copilot\footnote{\url{https://github.com/features/copilot/}} provide students with new methods for problem-solving and seeking learning resources. In the area of programming education, an increasing number of researchers are exploring the use of LLMs as teaching assistants to aid learners in studying programming languages or solving problems encountered in programming \cite{Savelka_2023, Kazemitabaar2023StudyingTE}.

To leverage LLM technology in aiding programming learners, Harvard University's CS50.AI is exploring the training of proprietary LLM like OpenAI's GPT-3.5 and GPT-4 for subject-specific educational assistance \cite{hicke2023aita}. Additionally, existing system \footnote{https://edition.cnn.com/2023/08/21/tech/khan-academy-ai-tutor/index.html} uses GPT-4 as an underlying functional model to answer learners' questions. Since direct using open-domain model tools involves sending queries to third-party administrators, there are data privacy concerns. To address the issue of user data leakage, recent works, like Hicke et al. \cite{hicke2023aita} has designed a training framework using open-source LLMs for programming education Q\&A tasks. This framework was trained on an English-language Q\&A dataset for entry-level CS courses, covering both Matlab and C programming languages.

With the booming development of the AI technology, an increasing number of people are learning the Python programming language. Unlike more specialized programming languages like Matlab and C, the learners of Python are not limited to students in computer-related fields. The learners come from diverse educational backgrounds, encompassing practitioners, AI researchers, and professionals in non-technical positions such as product management, operations, and finance. This diversity background gives research into Python programming instruction broader significance and applicability. Especially, for native Chinese-speaking Python learners, there is a noticeable lack of related research. Although AI tools like ChatGPT perform well in answering programming queries, their performance in responding to Python-related questions still exhibits some common sense errors, highlighting the importance of researching such AI tools.

\begin{figure}
    \centering
    \includegraphics[width=0.75\linewidth]{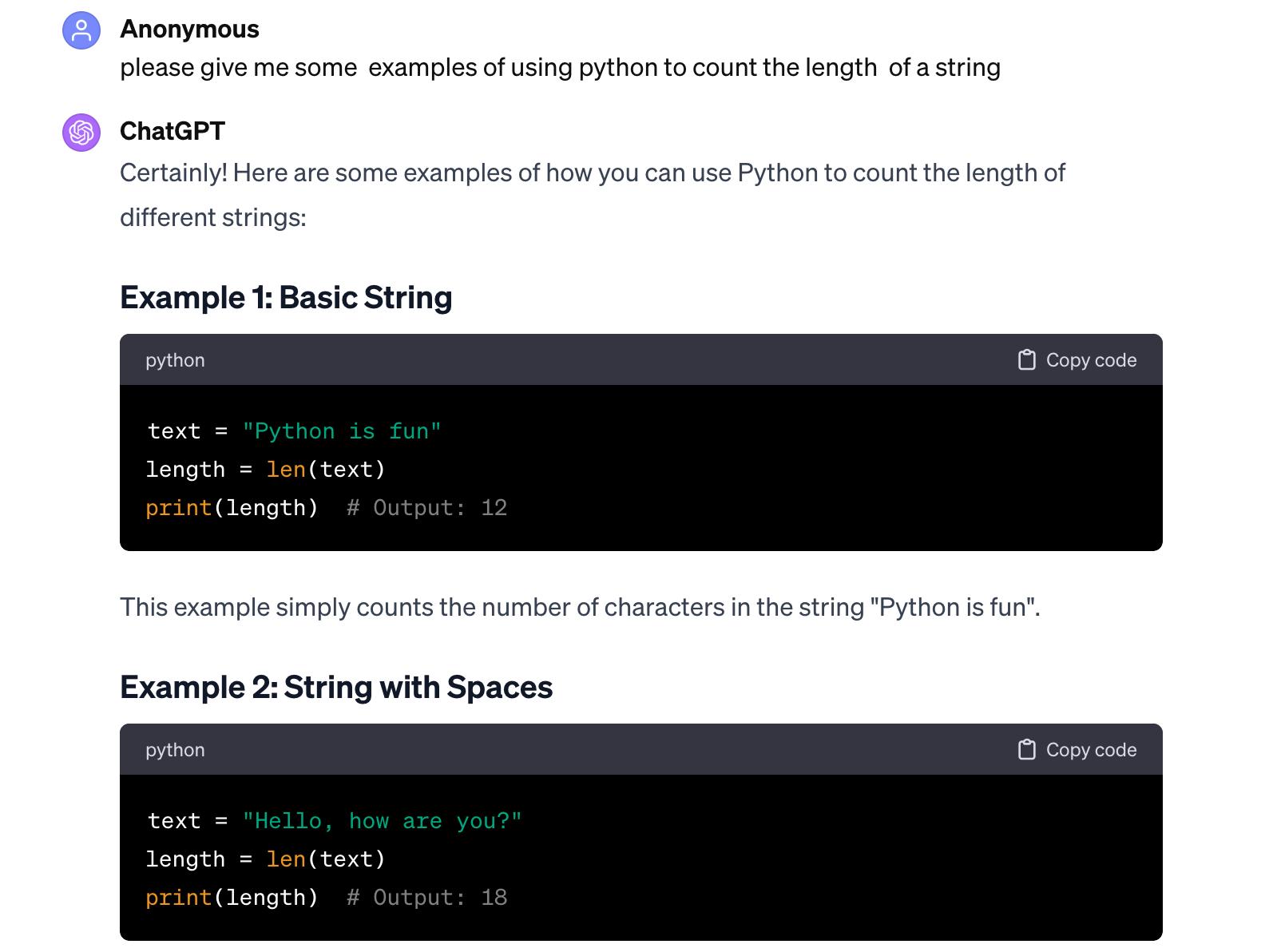}
    \caption{The example of GPT-4 in responding the Python-related questions (Test time is January 8, 2024).}
    \label{fig:intro}
\end{figure}

As shown in Fig \ref{fig:intro}, we ask GPT-4 about how to use python to count the length of String type variables. In the given example, the example codes are correct, but the results are wrong. The right string length of example 1 is 13, another is 19. \footnote{It is possible that this issue occurs randomly, and it may be resolved with different GPT versions. However, as a programming teaching assistant, this issue should be avoided as much as possible.} Such errors are hard to tolerate, because they can lead to learners losing confidence in using AI assistants for answering questions. Existing researches \cite{kiesler2023large, Doughty_2024} have pointed out that, although some errors are not too prevalent and may only occur occasionally, this issue is serious and requires human intervention for resolution. Therefore, it becomes particularly important to develop specialized, vertical LLMs for the programming education field to provide more accurate and professional programming guidance. 

It's important to note that whether it's providing AI teaching assistants for Python learners or further training specialized AI models in the field of programming education, both rely on the support of a large amount of high-quality data. In existing related works, datasets for diverse Python programming learners are still scarce. To address this challenge, this paper introduces a specialized Chinese single-turn question-and-answer (Q\&A) dataset for Python learners. This dataset contains 10,960 questions posed by Python learners at different learning stages and types. Based on user profiles, we annotated the answers to these questions from three dimensions: accessible answers to the question, classical analogies, and code examples. These dimensions are designed to help Python learners understand the questions they are confused about.

Besides, to ensure our answers provide additional educational value while meeting learners' basic Q\&A needs and are easy to understand and accept, we have referred to various educational theories like Bloom's Taxonomy of Educational Objectives \cite{article1}, Cognitive Load Theory \cite{LEPPINK2017385}, and Affective Education Theory \cite{article2}, combined with years of teaching experience, to carefully design answers for each question. The specific process of constructing answers is as follows:
\begin{enumerate}
    
    \item Design educational objectives for each question, determining the answer's approach and framework
    \item Based on the learners' characteristics and educational objectives, choose appropriate content and explanation methods (detailed explanation or in-depth analysis) to build an initial draft of the answer.
    \item Integrate affective teaching content, using encouraging language and positive feedback in the answers to enhance learners' confidence and self-efficacy.
\end{enumerate}

Additionally, to explore the potential of existing LLMs as AI teaching assistants in programming, we designed three benchmark testing tasks for our proposed dataset. The first task is the correctness of answer explanations, used to measure the accuracy of teaching content and assess the professional content reasoning abilities of general LLMs; the second is the consistency of questions and answers, used to measure whether LLMs can accurately understand diverse questions posed by learners; and the third is the usefulness of examples, used to assess the understanding of Chinese content by LLMs.

The main contributions of our research are summarized as follows:
\begin{itemize}
    \item We have proposed a new Chinese single-turn Q\&A dataset targeted at Python learners.

    \item For the constructed dataset, we have introduced three benchmark tasks for Python programming learning Q\&A, further exploring the capabilities of general LLMs in responding to professional programming knowledge and teaching ability.

    \item Experiments have shown that whether it is providing high-quality answers or accurately executing benchmark testing tasks, both present certain challenges for evaluated LLMs.
\end{itemize}

\section{Related Work}
\label{sec:related-word}

\subsection{Enhancing LLMs' Generation Capabilities with Data Support}
Large language models (LLMs) are typically trained on massive datasets, acquiring a broad range of language understanding and generative capabilities. Although pre-training provides a solid foundation, aligning LLMs with human knowledge is often necessary to achieve optimal performance in specific tasks. Thus, aligning LLMs with professional human knowledge is a key technique for effectively leveraging their pre-trained capabilities across a wide range of language tasks \cite{wang2023data, touvron2023llama}. This alignment not only enhances the models' adaptability but also enables a more precise understanding and response to human language and needs.

Improving the diverse task performance of LLMs through fine-tuning with human data has been extensively researched \cite{touvron2023llama, mialon2023augmented}. ClinicalGPT \cite{wang2023clinicalgpt} fine-tuned Bloom-7B in the healthcare domain using knowledge-base retrieval enhancements and human feedback-based reinforcement learning techniques, showing better performance in Q\&A tasks compared to baseline LLMs like LLaMA-7B and ChatGLM-6B assessed with GPT-4. Addressing educational questions in the medical field, Manathunga \cite{manathunga2023retrieval} leveraged expert knowledge to aid LLMs generation by retrieving and summarizing from large unstructured medical knowledge data, reducing the hallucination issues and potential harmful responses of the LLMs. Li \cite{li2023external} utilized the content generation ability of LLMs, integrating external semantic dictionary data to disambiguate homophones using LLMs. Therefore, high-quality professional knowledge data becomes crucial for empowering LLMs with more accurate and specialized content generation capabilities.

\subsection{LLMs as Programming Teaching Assistants}
With the development of research related to LLMs, more studies are evaluating the feasibility of LLMs in programming education \cite{zhong2023chatgpt, savelka2023large, arefin2023unmasking, kiesler2023exploring}, and attempting to train or further fine-tune specialized LLMs for the programming domain\footnote{https://sotblog.com/meet-the-khan-academy-chatbot-tutor-cnn-business/}. Anishka \cite{anishka2023chatgpt} explored the potential of using ChatGPT as a virtual teaching assistant (TA) in introductory programming courses. Although ChatGPT possesses strong content understanding capabilities, using it directly as a programming teaching assistant still requires supervision by experienced teaching assistants, especially for more complex programming-related questions. AI-TA \cite{hicke2023aita} combined retrieval-augmented generation (RAG) \cite{lewis2021retrievalaugmented}, supervised fine-tuning (SFT), and learning from human preference data using direct preference optimization (DPO) \cite{rafailov2023direct} to fine-tune LLaMA-2 \cite{touvron2023llama} with the Piazza dataset for introductory computer science courses, creating a Q\&A assistant relevant to programming education. However, this work's training data only included Matlab and C programming languages.

Therefore, research how to use LLMs as programming teaching assistants still has significant value. The primary challenge in this area is the insufficiency of existing data resources. Especially for Python learners with more diverse knowledge backgrounds, the corresponding Q\&A corpora is especially scarce. To address this problem, we established a Chinese Q\&A dataset for Python learners. To the best of our knowledge, this is the first Chinese Q\&A dataset for Python programming.

\begin{figure}
    \centering
    \includegraphics[width=1\linewidth]{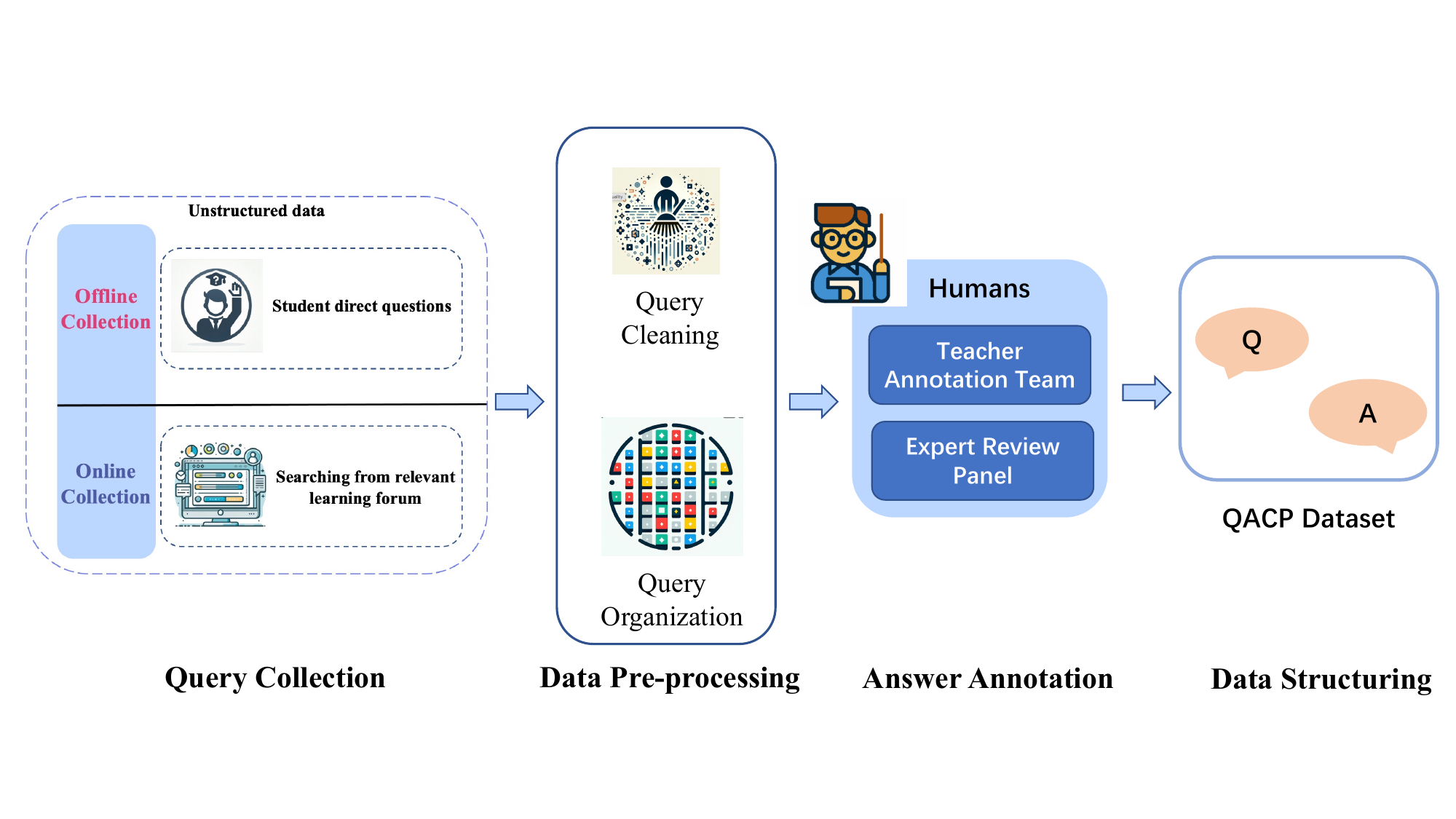}
    \caption{The data construction pipeline of the QACP dataset.}
    \label{fig:processing}
\end{figure}

\section{Dataset Construction}
\label{sec:evaluation-tasks}

In this section, we systematically introduce how to construct the dataset and data analysis to demonstrate the characteristics of the dataset. The overall data construction process is shown in Figure \ref{fig:processing}.

\subsection{Query Collection}
To ensure that the question design accurately reflects the real-world questioning patterns and requirements of Python learners, we implemented a comprehensive methodology. Firstly, we engaged in an extensive data collecting process aimed at learners from variously learning stages. These learners have diverse educational backgrounds and different proficiency levels in Python, guaranteeing that the collected questions encompass a wide range of complexities.

We gathered 50,247 Python-related questions from offline and online collcetions. These questions cover all knowledge points of Python course, ranging from basic concepts, Python syntax to complex programming issues, and from understanding data structures to the application of algorithms and other aspects. Subsequently, we categorized and organized these questions. To ensure the quality and representativeness of the data, the selection process was carried out by experienced Python educators. This process involved classifying, filtering, and restructuring the questions to ensure that each one has a clear expression, specific learner background, and a definitive question description. We excluded repetitive, ambiguous, or questions not directly related to Python learning to ensure that the final selection of questions is both high-quality and highly relevant.

\begin{figure}
    \centering
    \includegraphics[width=0.5\linewidth]{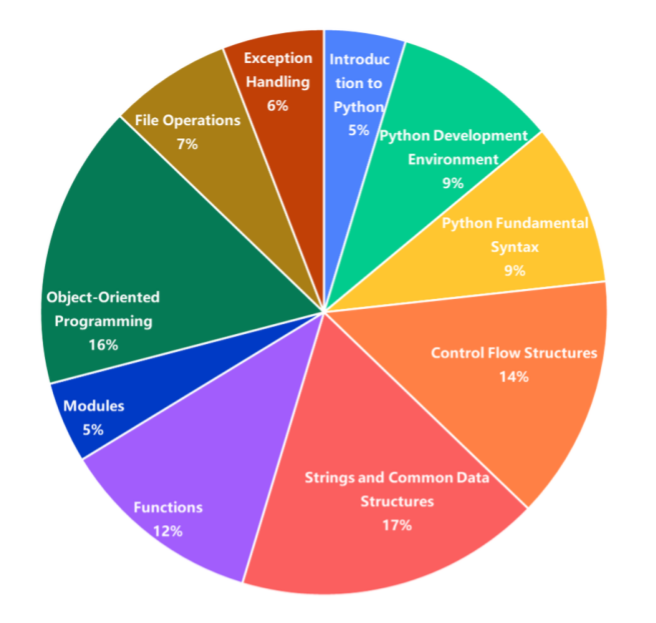}
    \caption{The distribution of knowledge points for the collected Python questions.}
    \label{fig:distribution}
\end{figure}

In the end, we marked answers for 10960 questions. We aim to build a realistic, practical, and efficient Python learning Q\&A database that helps learners better learn the Python programming language and address problems encountered in actual programming. The distribution of knowledge points for the collected Python questions is shown in Figure \ref{fig:distribution}.

\subsection{Annotation Principle and Quality Control}
Given that the dataset we are developing is focused on single-question-and-answer pairs, specifically designed to address the queries of Python learners. Concerning their specific learning challenges, we necessitate annotations from three key perspectives: accessible answers to the question, classic analogies, and practical code examples.

To ensure the high quality of the annotated answers, we implemented a rigorous quality control process. Considering the need for profound domain knowledge in Python programming and substantial teaching experience for the creation of this dataset, we selectively recruited 16 professional instructors specializing in Python programming as our annotation experts. Prior to commencing the official annotation process, these experts underwent thorough training. This training was designed to ensure they not only fully grasped our specific annotation requirements but also became proficient in utilizing the annotation system we developed.

Throughout the annotation process, we implemented a dual-verification mechanism to guarantee the accuracy and dependability of the results. In this system, every annotation underwent two separate rounds of verification post-submission. Initially, an annotation expert conducted the first round, meticulously examining each submission, and suggesting revisions for any errors or ambiguities identified. After implementing these corrections, a second expert embarked on the subsequent verification phase. This stage involved a thorough review of annotations that had successfully passed the initial round, to ensure compliance with our stringent standards. Annotations that were validated as accurate in this second phase were then officially approved.

Through this hierarchical, multi-stage quality control program, we can ensure that every answer in the dataset has been rigorously reviewed, thus guaranteeing answers accuracy and professionalism. This not only helps to enhance the overall quality of the dataset but also provides a reliable and efficient learning resource for Python learners.

To ensure the consistency of annotations, after annotating a certain amount of data, we randomly sampled 1,000 Q\&A pair for consistency verification. Two annotation experts independently scored these annotations for consistency, calculating the overall quality of the answers. The scores were then used to compute the kappa value for annotation consistency. The final score was 0.89 (The score closer to 1 indicates higher consistency), demonstrating that our annotations are highly consistent, thereby ensuring the objectivity of the answers to the greatest extent possible.

\subsection{Data Analysis}
Table \ref{tab1} shows the overall statistical data of the annotated dataset. The longest question for Beginners is 51 tokens, while for Experienced learners it's slightly longer at 58 tokens. The combined longest question length is an average of 54.5 tokens. The shortest question length is 5 tokens for Beginners and 7 for Experienced learners, with an overall shortest length of 6 tokens. On average, questions asked by Beginners are 18.29 tokens long, those by Experienced learners are 18.02 tokens, and the average length across all questions is 18.15 tokens. For the answer, we dissected into two answer types: "Explain in Detail" and "In-depth Explanation." For detailed explanations, the maximum number of tokens per answer is 1,290, the minimum is 144 tokens, and the average is 485.41 tokens. In-depth explanation answers are longer than Explain in Detail. The answers with code annotations account for about 70\% of the overall annotated answers, indicating that our dataset not only explains the concepts but also provides concrete code examples to help learners study Python.% As shown in this Table, our annotated dataset indicates a balance of question length across experience levels and a more considerable variation in answer lengths and content.    

\begin{table}
    \centering

    \small
    \begin{tabular}{c|p{3cm}<{\centering}|p{3cm}<{\centering}|c}
    \hline
         \multirow{2}{*}{Data Type} & \multicolumn{2}{c|}{Question Type} & Total/Avg\\
         \cline{2-4} 
         &\makecell{Beginner}&  \makecell{Experience in Programming} & \\ 
         \hline
         Number of dialogues &  7712 & 3248 & 10960\\
        \hline
         Longest question length & 51 & 58 & 54.5\\
         Shortest question length & 5 & 7 & 6\\
         Average question length & 18.29 & 18.02 & 18.15\\
        \hline
            &  \multicolumn{2}{c|}{Answer Type} & \\
        \cline{2-3} 
        & Explain in Detail & In-depth Explanation & \\ 
        \hline
         Max. \# tokens per answer & 1290 & 1323 & 1306.5 \\
         Min. \# tokens per answer & 144 & 350 & 247\\
         Avg. \# tokens per answer & 485.41  & 532.19 & 508.08\\
         \hline
         Number of answers containing code  & 5016 & 2107 & 7123\\
    \hline
    \end{tabular}
    \caption{Overall statistics of the QACP dataset.}
    \label{tab1}
\end{table}

\section{Experiment}
\label{sec:exper}

To further concretely assess the Chinese response capabilities of LLMs for Python-related questions, we set up the following experimental framework.

\subsection{Benchmark Models}
In our experiments, we tested a series of LLMs known for their proficiency in Chinese understanding and generation capabilities. To comprehensively evaluate the abilities of these LLMs, including the impact of model size on content understanding and answer generation, we used a variety of models with different parameter sizes. The specific benchmark models used are as follows:

GPT-3.5-turbo\footnote{\url{https://platform.openai.com/docs/models/}}: Developed by OpenAI, GPT-3.5 is a language model capable of a wide range of natural language processing tasks such as translation, summarization, Q\&A, and text completion. It generates coherent responses relevant to given inputs.

GPT-4: An upgrade of GPT-3.5, GPT-4 not only maintains the powerful natural language processing abilities of its predecessor but also has significant improvements in model architecture, algorithm optimization, and data handling. It can more effectively understand complex text structures, contextual meanings, and user intentions, thus providing more accurate and rich responses.

Gemini-1.0 \footnote{\url{https://gemini.google.com/}}: Developed by Google, Gemini 1.0 can generalize and seamlessly understand, operate across, and combine different types of information, including text, code, audio, images, and videos. Gemini 1.0 is optimized for different sizes: Ultra, Pro, and Nano. In our experiments, we used Gemini-1.0-Pro.

Baichuan \cite{yang2023baichuan}: Baichuan is an open-source, large-scale multilingual language model with 13 billion parameters, trained from scratch on 2.6 trillion tokens. It excels in conversation and contextual understanding. We used Baichuan2-turbo, Baichuan-13B in our experiments.

ChatGLM \cite{zeng2023glm130b}: An open-source bilingual model based on a general language model framework. ChatGLM has undergone specific optimizations including supervised fine-tuning, feedback-guided, and reinforcement learning combined with human feedback. We used ChatGLM-6B, ChatGLM-turbo and ChatGLM-4 in our experiments.

Qwen \cite{bai2023qwen}: A comprehensive language model series including models with varying parameter sizes. The foundational models consistently show excellent performance in numerous downstream tasks. We used Qwen-14B in our experiments.

Moonshot\footnote{\url{https://www.moonshot.cn/}}: Developed by Moonshot AI Technologies, Moonshot is a large language model focusing on processing and understanding long text data, capable of supporting inputs up to 200,000 characters.

LLaMA-2-chat \cite{touvron2023llama}: The second generation of Meta's open-source large language model series. Built on LLaMA, it increases the number of training tokens and extends the context length to 4096. In our experiments, we used LLaMA-2-chat-7B, Atom-7B, LLaMA-2-chat-13B\footnote{As LLaMA-2-chat-70B does not support Chinese, this model was not selected.}.

\subsection{Data Sampling}
The experiments aims to evaluate the performance of existing general LLMs in generating answers to Python programming questions and to deeply analyze their understanding and generative capabilities. To this end, we carefully selected 534 Q\&A examples from the constructed dataset for experimentation. These examples cover common knowledge points of Python course and were chosen through random sampling to ensure the diversity and comprehensiveness, thereby providing a thorough evaluation of the LLMs' performance. \footnote{Sampling for annotation is conducted according to the distribution proportions of 10 modules in the total sample. This involves drawing samples from each module such that the proportion of the sample reflects the composition of the full dataset. The questions selected for sampling are typically common ones within each module, covering various subtopics of knowledge. This approach ensures that the sample is representative of the overall data set in terms of content and distribution across different areas of knowledge.} Through this approach, we ensured that the selected examples not only reflect a range of typical problems in Python programming but also provide a comprehensive assessment of the LLMs' capabilities in answering various types of programming questions.

\subsection{Evaluation Principles}
In the scenarios of educational assistance, the practicality and accuracy of responses are crucial evaluation criteria. Previous research \cite{shi2023sotana} has indicated that traditional evaluation metrics like BLEU \cite{papineni2002bleu}, Rouge-L \cite{lin2004rouge}, and Cider \cite{vedantam2015cider} are not effective in capturing the practicality and accuracy of model responses in generative tasks. To measure the existing general LLMs' ability to generate answers for Python programming-related questions and to further analyze their Python knowledge reasoning abilities, we use both human-based and LLMs-based evaluations.

\textbf{Human Evaluation of LLMs Answer Generation: }In this experiment, we employ human evaluations to assess the overall quality of answers generated by LLMs. To ensure that the answer generation by LLMs conforms to our manually constructed annotation principles, we conduct a comprehensive scoring evaluation from three aspects: the correctness of answer explanations, the consistency of questions and answers, and the usefulness of answers, followed by calculating the overall score (the higher the score, the better the quality). In this process, we give the same prompts and questions to different LLMs for generating responses, and finally, we use human evaluation to judge the results generated by the LLMs.

The specific quantitative metric design principles for the generated results are as follows:
\begin{enumerate}
    \item We quantify the above three aspects of answer generation into the following four evaluation dimensions: Technical accuracy and rigor of answer (Answer Correctness, AC), Logical organization and expressiveness of language (Expressive Ability, EA), Whether the content of the answer solves the problem (Usefulness, UF), and Whether it aligns with the type of learner (Consistency with Learner, CL). 
    
    \item The final evaluation metric is a comprehensive satisfaction index: Excellent, Good, Average, Poor (out of 100; Excellent: 90-100; Good: 80-90; Average: 60-70; Poor: below 60). 
    
    \item For questions that require code generation, we score the accuracy of the generated code separately (1 for correct, 0 for incorrect), and finally report the accuracy rate of code generation.
\end{enumerate}

The weight distribution standards for different dimensions are shown in Table \ref{tab:metrics}.
And the calculation result of the final composite score is shown in Equation (1).

\begin{table}
    \centering
    \large
    \begin{tabular}{c|cccc}
    \hline
         Evaluation Dimension &  AC &  EA & UF &  CL \\
         \hline
         Weights &  60\% &  5\%&  20\%& 15\%\\
    \hline
    \end{tabular}
    \caption{Metric principle. AC (Answer Correctness) means to metric technical accuracy and rigor of answer. EA (Expressive Ability) means the logical organization and expressiveness of language. UF (Usefulness) means whether the content of the answer solves the problem. CL (Consistency with Learner) means whether it aligns with the type of learner.} 
    \label{tab:metrics}
\end{table}

\begin{equation}
    \emph{comprehensive satisfaction index} \&= 0.6 * AC + 0.05 * EA + 0.2 * UF + 0.15 * CL   
\end{equation}    

To ensure fairness in our experiments, we maintained a consistent prompt for different LLMs. The prompt used in the experiment was: "You are a Python teacher, and I am a Python learner, please answer my question:."

\textbf{Evaluation of LLMs' Reasoning Ability in Python Q\&A}: To more specifically analyze the LLMs' ability to generate answers for Python programming questions, we provided both the answers and the annotated questions to the LLMs. Based on the three human evaluation perspectives mentioned above, we designed three benchmark tasks along with corresponding evaluation standards:

\begin{enumerate}
    \item \textbf{Correctness of Answer Explanation}: This is used to assess the accuracy of the LLM's teaching content explanation and to test the model's code checking ability. The main focus is to evaluate the model's capacity to detect professional relevance errors and illusions. Since the correctness of the response is crucial for answering questions, we divided this metric into two categories: incorrect explanation and correct explanation. To further assess the LLMs' code checking ability, we also separately calculated the accuracy of compilation for questions requiring code examples. 

    \item \textbf{Consistency Between Question and Answer}: This measures whether the LLM can accurately understand the learner's query. For this task, we assigned three specific scores for the LLMs to rate: 0.5 (weakly relevant), 1 (relevant), 1.5 (strongly relevant).

    \item \textbf{Helpfulness}: This is used to measure the LLMs' ability to understand professional content. We used two scores, inappropriate and appropriate, for the LLMs to evaluate.
\end{enumerate}

To validate the context comprehension ability of LLMs, we randomly chose 30\% of the samples and performed error rewriting in the answers, i.e., marking some key content as incorrect, to test the detection ability of the LLMs used for evaluation. \emph{It is important to note that this part of the experiment is set up to analyze the reasoning capabilities of LLMs.} Therefore, we only used the best-performing model in the same category from the aforementioned benchmark models for evaluation. The aim is to explore how LLMs' ability to answer Python questions can be improved.

\section{Experiment Results and Analysis}

\subsection{Python Problem-Solving Ability Evaluation of LLMs}

\begin{table}
    \centering
    \begin{tabular}{c|ccccc|c}
    \hline
        LLMs & AC & EA & UF & CL & Overall Score & Code Accuracy (\%)\\
        \hline
        GPT-4 & \textbf{78.46} & 75.82  & 67.95  & \textbf{65.57} & \textbf{74.62} &  \textbf{96.29} \\
        GPT-3.5-turbo & 72.03 & 75.29  & 67.22  & 64.82 & 70.63 & 92.3\\
        Gemini-1.0-pro & 58.47 & 59.8 & 50.32  &  48.26 & 55.49 & 85\\
        ChatGLM4 & 73.91 & 76.91 & \textbf{68.85} & 63.49 & 71.02 & 92.5\\
        ChatGLM-turbo & 70.5  & \textbf{79.1} & 56.4 & 54 & 65.6 & 95\\
        Baichuan2-turbo & 68.25 & 66.56 & 60.30 & 57.20 & 64.92 & 80.43 \\ 
        Qwen-14B & 58.25 & 73.2 & 61.04  &  56.4& 59.89 & 79\\
        Baichuan-13B & 55.22 & 56.94 & 45.89 & 39.54 & 51.09 &61.82 \\
        Moonshot-v1-32k & 69.48 & 69.63 & 59.96  & 57.34 & 66 & 87.93\\
        LLaMA-2-chat-13B & 42.06 & 65.03 & 37.60 & 37.04 & 41.65  & 45.45\\ 
        ChatGLM-6B &  50.02  & 75.64  & 39.79  & 37.60 &  46.19& 65.52\\
        LLaMA-2-chat-7B &  5.37 & 49.00 & 5.30 & 5.30 & 9.07 & 0.00\\
        Atom-7B & 24.40 & 50.00 & 15.50 & 14.40 & 22.40 & 0.00\\ 
        
    \hline
    \end{tabular}
    \caption{Python problem solving ability evaluation results of LLMs. Overall Score means the comprehensive satisfaction index, which is computed by Eq(1). AC means technical accuracy and rigor of answer, EA indicates expressive ability, UF indicates the answer can solve the problem, and CL means the answer aligns with the type of learner. Code Accuracy means the accuracy rate of Python code generation.}
    \label{tab:my_label}
\end{table}
Based on the manual scoring criteria established in Table \ref{sec:evaluation-tasks}, and considering the issues extracted from our constructed dataset, we report the manual scoring results of Language Learning Models (LLMs) in answering Python questions in Table \ref{tab:my_label}. We specifically hired teachers with Python teaching experience to grade the outputs of LLMs. From the results in Table \ref{tab:my_label}, it can be seen that, in terms of overall satisfaction, GPT-4 still performs the best, indicating that GPT-4 is far superior to other LLMs in understanding and generating answers to Chinese questions. However, there are still some issues with its expressive ability. Some Chinese LLMs, such as ChatGLM-turbo, have caught up with or even surpassed GPT-4 in terms of fluency in answering Python-related questions in Chinese. Especially for the newest GLM model (Chat-GLM4), the value of usefulness surpasses that of GPT-4. 

Besides, based on the results shown in the table, we can see that the size of the model parameters remains a primary factor affecting model performance. The larger the number of parameters in a model, the more it improves in terms of understanding and answering questions.

\subsection{Python Answer Reasoning Capability Analysis of LLMs}

In order to further analyze the LLMs' ability on specialized content about Python, we evaluated them from three perspectives: accuracy of knowledge points, QA consistency, and usefulness of examples. This experiment allowed us to assess their capability in understanding and applying Python content, comprehending user intentions, and grasping Chinese content. To further investigate the ability of LLMs to utilize knowledge in vertical domains, we randomly extracted 30\% of the answers to generate adversarial samples. For these answers, we conducted error-prone rewriting, altering some codes and explanations of professional content to incorrect ones. The experimental results are presented in Table \ref{tab:res2}. Based on the results of this Table, we can conclude :

\begin{table}
    \centering
    \begin{tabular}{c|cccc}
    \hline
        LLMs & Explanation Accuracy & Code Accuracy &  Consistency & Helpfulness \\
        \hline
        GPT-4 & 85.4 & 96.1 & 99.1 & 98.7\\
        GPT-3.5-turbo & 79.4 & 84.8 & 98.3 & 98.1  \\
        Gemini-1.0-pro & 80.14 & 83.7 & 99.25 & 98.5 \\
        ChatGLM-turbo & 79.6 & 84.3 & 99.1 & 100  \\
        ChatGLM4 & 88.02 & 93.07 & 98.88 & 97.37 \\
        Baichuan2-turbo & 82.4 & 88.9 & 96.4 & 99.4 \\
        Moonshot-v1-32k & 82.95 & 89.7 & 96.62 & 95.5 \\
        \hline
        Human & 100 & 100 & 100 & 100 \\
    \hline
    \end{tabular}
    \caption{Python answer reasoning capability analysis results of LLMs.}
    \label{tab:res2}
\end{table}

\begin{enumerate}
    \item For general large language models, even GPT-4, are not perform well when it comes to comprehending Python-related specialized knowledge. As shown by the accuracy of professional content explanations, the LLMs used for evaluation do not achieve a high accuracy rate in error detection (Even GPT-4 only obtains 85.4\%).
    
    \item Chinese LLMs do not perform well in understanding code logic when we introduce errors into the code, revealing that general LLMs are prone to producing professionally incorrect answers when addressing specialized questions. Such errors are unacceptable in educational applications. Compared to other models used for evaluation, GPT-4 has a higher accuracy rate in detecting code errors.
    
    \item In terms of understanding contextual relationships, Chinese LLMs demonstrate its formidable capabilities. As shown by Consistency and Helpfulness, ChatGLM performs better than GPT-4. This indicates that existing Chinese open-source LLM frameworks have the feasibility and application potential on developing specialized vertical-domain LLMs for Chinese application scenarios.
\end{enumerate}

\section{Conclusion and Further Work}
\label{sec:Conclusion}

In this study, we developed QACP, a question-and-answer dataset specifically designed for Chinese learners who want to study Python. We collected a large number of questions from Python learners with differently learning backgrounds, and then systematically categorized and organized these questions. Finally, we compiled 10,960 questions, providing detailed annotations and categorizations for each question. To demonstrate the value of our dataset and further test the capability of existing large language models in answering Python questions in Chinese, we design two experiments to evaluate these models' abilities in answer generation and knowledge understanding. According to our results, the evaluated LLMs achieve poor performance on all experiment, reflecting the challenge and practical value of the dataset.

In future research, we plan to continue designing experiments to evaluate more newly LLMs and further analyze their abilities in answering programming questions. Additionally, we aim to explore how to fully leverage our constructed dataset, aim to develop a specialized Python teaching assistant for Chinese learners. We hope to create an intelligent teaching assistant system that can provide a personalized learning experience, not only answering programming-related questions but also offering customized learning suggestions. Furthermore, we intend to further expand the dataset to accommodate a more diverse range of learner backgrounds. 

% \section*{Acknowledgments}

\bibliography{references.bib}

\begin{thebibliography}{30}
\providecommand{\natexlab}[1]{#1}
\providecommand{\url}[1]{\texttt{#1}}
\expandafter\ifx\csname urlstyle\endcsname\relax
  \providecommand{\doi}[1]{doi: #1}\else
  \providecommand{\doi}{doi: \begingroup \urlstyle{rm}\Url}\fi

\bibitem[Anishka et~al.(2023)Anishka, Mehta, Gupta, Kumar, and Jalote]{anishka2023chatgpt}
Anishka, A.~Mehta, N.~Gupta, D.~Kumar, and P.~Jalote.
\newblock Can chatgpt play the role of a teaching assistant in an introductory programming course?, 2023.

\bibitem[Arefin et~al.(2023)Arefin, Heya, Al-Qudah, Ineza, and Serwadda]{arefin2023unmasking}
S.~E. Arefin, T.~A. Heya, H.~Al-Qudah, Y.~Ineza, and A.~Serwadda.
\newblock Unmasking the giant: A comprehensive evaluation of chatgpt's proficiency in coding algorithms and data structures, 2023.

\bibitem[Bai et~al.(2023)Bai, Bai, Chu, Cui, Dang, Deng, Fan, Ge, Han, Huang, Hui, Ji, Li, Lin, Lin, Liu, Liu, Lu, Lu, Ma, Men, Ren, Ren, Tan, Tan, Tu, Wang, Wang, Wang, Wu, Xu, Xu, Yang, Yang, Yang, Yang, Yao, Yu, Yuan, Yuan, Zhang, Zhang, Zhang, Zhang, Zhou, Zhou, Zhou, and Zhu]{bai2023qwen}
J.~Bai, S.~Bai, Y.~Chu, Z.~Cui, K.~Dang, X.~Deng, Y.~Fan, W.~Ge, Y.~Han, F.~Huang, B.~Hui, L.~Ji, M.~Li, J.~Lin, R.~Lin, D.~Liu, G.~Liu, C.~Lu, K.~Lu, J.~Ma, R.~Men, X.~Ren, X.~Ren, C.~Tan, S.~Tan, J.~Tu, P.~Wang, S.~Wang, W.~Wang, S.~Wu, B.~Xu, J.~Xu, A.~Yang, H.~Yang, J.~Yang, S.~Yang, Y.~Yao, B.~Yu, H.~Yuan, Z.~Yuan, J.~Zhang, X.~Zhang, Y.~Zhang, Z.~Zhang, C.~Zhou, J.~Zhou, X.~Zhou, and T.~Zhu.
\newblock Qwen technical report, 2023.

\bibitem[Doughty et~al.(2024)Doughty, Wan, Bompelli, Qayum, Wang, Zhang, Zheng, Doyle, Sridhar, Agarwal, Bogart, Keylor, Kultur, Savelka, and Sakr]{Doughty_2024}
J.~Doughty, Z.~Wan, A.~Bompelli, J.~Qayum, T.~Wang, J.~Zhang, Y.~Zheng, A.~Doyle, P.~Sridhar, A.~Agarwal, C.~Bogart, E.~Keylor, C.~Kultur, J.~Savelka, and M.~Sakr.
\newblock A comparative study of ai-generated (gpt-4) and human-crafted mcqs in programming education.
\newblock In \emph{Proceedings of the 26th Australasian Computing Education Conference}, ACE 2024. ACM, Jan. 2024.
\newblock \doi{10.1145/3636243.3636256}.
\newblock URL \url{http://dx.doi.org/10.1145/3636243.3636256}.

\bibitem[Hicke et~al.(2023)Hicke, Agarwal, Ma, and Denny]{hicke2023aita}
Y.~Hicke, A.~Agarwal, Q.~Ma, and P.~Denny.
\newblock Ai-ta: Towards an intelligent question-answer teaching assistant using open-source llms, 2023.

\bibitem[Katz and Romi(2003)]{article2}
Y.~Katz and S.~Romi.
\newblock Affective education: The nature and characteristics of teachers and students attitudes toward school.
\newblock \emph{Educational Practice and Theory}, 25:\penalty0 35--47, 01 2003.
\newblock \doi{10.7459/ept/25.1.04}.

\bibitem[Kazemitabaar et~al.(2023)Kazemitabaar, Chow, Ma, Ericson, Weintrop, and Grossman]{Kazemitabaar2023StudyingTE}
M.~Kazemitabaar, J.~T. Chow, C.~K.~T. Ma, B.~Ericson, D.~Weintrop, and T.~Grossman.
\newblock Studying the effect of ai code generators on supporting novice learners in introductory programming.
\newblock \emph{Proceedings of the 2023 CHI Conference on Human Factors in Computing Systems}, 2023.
\newblock URL \url{https://api.semanticscholar.org/CorpusID:256868626}.

\bibitem[Kiesler and Schiffner(2023)]{kiesler2023large}
N.~Kiesler and D.~Schiffner.
\newblock Large language models in introductory programming education: Chatgpt's performance and implications for assessments, 2023.

\bibitem[Kiesler et~al.(2023)Kiesler, Lohr, and Keuning]{kiesler2023exploring}
N.~Kiesler, D.~Lohr, and H.~Keuning.
\newblock Exploring the potential of large language models to generate formative programming feedback, 2023.

\bibitem[Lee et~al.(2023)Lee, Myung, Han, Jin, and Oh]{lee2023learning}
C.~Lee, J.~Myung, J.~Han, J.~Jin, and A.~Oh.
\newblock Learning from teaching assistants to program with subgoals: Exploring the potential for ai teaching assistants, 2023.

\bibitem[Leppink(2017)]{LEPPINK2017385}
J.~Leppink.
\newblock Cognitive load theory: Practical implications and an important challenge.
\newblock \emph{Journal of Taibah University Medical Sciences}, 12\penalty0 (5):\penalty0 385--391, 2017.
\newblock ISSN 1658-3612.
\newblock \doi{https://doi.org/10.1016/j.jtumed.2017.05.003}.
\newblock URL \url{https://www.sciencedirect.com/science/article/pii/S1658361217300835}.

\bibitem[Lewis et~al.(2021)Lewis, Perez, Piktus, Petroni, Karpukhin, Goyal, Küttler, Lewis, tau Yih, Rocktäschel, Riedel, and Kiela]{lewis2021retrievalaugmented}
P.~Lewis, E.~Perez, A.~Piktus, F.~Petroni, V.~Karpukhin, N.~Goyal, H.~Küttler, M.~Lewis, W.~tau Yih, T.~Rocktäschel, S.~Riedel, and D.~Kiela.
\newblock Retrieval-augmented generation for knowledge-intensive nlp tasks, 2021.

\bibitem[Li(2023)]{li2023external}
C.~Li.
\newblock External knowledge augmented polyphone disambiguation using large language model, 2023.

\bibitem[Lin(2004)]{lin2004rouge}
C.-Y. Lin.
\newblock Rouge: A package for automatic evaluation of summaries.
\newblock In \emph{Text summarization branches out}, pages 74--81, 2004.

\bibitem[Manathunga and Illangasekara(2023)]{manathunga2023retrieval}
S.~S. Manathunga and Y.~A. Illangasekara.
\newblock Retrieval augmented generation and representative vector summarization for large unstructured textual data in medical education, 2023.

\bibitem[Mialon et~al.(2023)Mialon, Dessì, Lomeli, Nalmpantis, Pasunuru, Raileanu, Rozière, Schick, Dwivedi-Yu, Celikyilmaz, Grave, LeCun, and Scialom]{mialon2023augmented}
G.~Mialon, R.~Dessì, M.~Lomeli, C.~Nalmpantis, R.~Pasunuru, R.~Raileanu, B.~Rozière, T.~Schick, J.~Dwivedi-Yu, A.~Celikyilmaz, E.~Grave, Y.~LeCun, and T.~Scialom.
\newblock Augmented language models: a survey, 2023.

\bibitem[Mizbani and Chalak(2017)]{article1}
M.~Mizbani and A.~Chalak.
\newblock Analyzing reading and writing activities of iranian efl textbook prospect 3 based on bloom's revised taxonomy.
\newblock 4:\penalty0 13--27, 01 2017.

\bibitem[{OpenAI}(2023)]{openai2023chatgpt}
{OpenAI}.
\newblock Chatgpt: Optimizing language models for dialogue, 2023.
\newblock URL \url{https://openai.com/chatgpt}.
\newblock Software available from OpenAI.

\bibitem[Papineni et~al.(2002)Papineni, Roukos, Ward, and Zhu]{papineni2002bleu}
K.~Papineni, S.~Roukos, T.~Ward, and W.-J. Zhu.
\newblock Bleu: a method for automatic evaluation of machine translation.
\newblock In \emph{Proceedings of the 40th annual meeting of the Association for Computational Linguistics}, pages 311--318, 2002.

\bibitem[Rafailov et~al.(2023)Rafailov, Sharma, Mitchell, Ermon, Manning, and Finn]{rafailov2023direct}
R.~Rafailov, A.~Sharma, E.~Mitchell, S.~Ermon, C.~D. Manning, and C.~Finn.
\newblock Direct preference optimization: Your language model is secretly a reward model, 2023.

\bibitem[Savelka et~al.(2023{\natexlab{a}})Savelka, Agarwal, An, Bogart, and Sakr]{Savelka_2023}
J.~Savelka, A.~Agarwal, M.~An, C.~Bogart, and M.~Sakr.
\newblock Thrilled by your progress! large language models (gpt-4) no longer struggle to pass assessments in higher education programming courses.
\newblock In \emph{Proceedings of the 2023 ACM Conference on International Computing Education Research V.1}, ICER 2023. ACM, Aug. 2023{\natexlab{a}}.
\newblock \doi{10.1145/3568813.3600142}.
\newblock URL \url{http://dx.doi.org/10.1145/3568813.3600142}.

\bibitem[Savelka et~al.(2023{\natexlab{b}})Savelka, Agarwal, Bogart, and Sakr]{savelka2023large}
J.~Savelka, A.~Agarwal, C.~Bogart, and M.~Sakr.
\newblock Large language models (gpt) struggle to answer multiple-choice questions about code, 2023{\natexlab{b}}.

\bibitem[Shi et~al.(2023)Shi, Zhang, Wang, Chen, Du, Zhang, Han, Zhang, and Sun]{shi2023sotana}
E.~Shi, F.~Zhang, Y.~Wang, B.~Chen, L.~Du, H.~Zhang, S.~Han, D.~Zhang, and H.~Sun.
\newblock Sotana: The open-source software development assistant, 2023.

\bibitem[Touvron et~al.(2023)Touvron, Martin, Stone, Albert, Almahairi, Babaei, Bashlykov, Batra, Bhargava, Bhosale, Bikel, Blecher, Ferrer, Chen, Cucurull, Esiobu, Fernandes, Fu, Fu, Fuller, Gao, Goswami, Goyal, Hartshorn, Hosseini, Hou, Inan, Kardas, Kerkez, Khabsa, Kloumann, Korenev, Koura, Lachaux, Lavril, Lee, Liskovich, Lu, Mao, Martinet, Mihaylov, Mishra, Molybog, Nie, Poulton, Reizenstein, Rungta, Saladi, Schelten, Silva, Smith, Subramanian, Tan, Tang, Taylor, Williams, Kuan, Xu, Yan, Zarov, Zhang, Fan, Kambadur, Narang, Rodriguez, Stojnic, Edunov, and Scialom]{touvron2023llama}
H.~Touvron, L.~Martin, K.~Stone, P.~Albert, A.~Almahairi, Y.~Babaei, N.~Bashlykov, S.~Batra, P.~Bhargava, S.~Bhosale, D.~Bikel, L.~Blecher, C.~C. Ferrer, M.~Chen, G.~Cucurull, D.~Esiobu, J.~Fernandes, J.~Fu, W.~Fu, B.~Fuller, C.~Gao, V.~Goswami, N.~Goyal, A.~Hartshorn, S.~Hosseini, R.~Hou, H.~Inan, M.~Kardas, V.~Kerkez, M.~Khabsa, I.~Kloumann, A.~Korenev, P.~S. Koura, M.-A. Lachaux, T.~Lavril, J.~Lee, D.~Liskovich, Y.~Lu, Y.~Mao, X.~Martinet, T.~Mihaylov, P.~Mishra, I.~Molybog, Y.~Nie, A.~Poulton, J.~Reizenstein, R.~Rungta, K.~Saladi, A.~Schelten, R.~Silva, E.~M. Smith, R.~Subramanian, X.~E. Tan, B.~Tang, R.~Taylor, A.~Williams, J.~X. Kuan, P.~Xu, Z.~Yan, I.~Zarov, Y.~Zhang, A.~Fan, M.~Kambadur, S.~Narang, A.~Rodriguez, R.~Stojnic, S.~Edunov, and T.~Scialom.
\newblock Llama 2: Open foundation and fine-tuned chat models, 2023.

\bibitem[Vedantam et~al.(2015)Vedantam, Lawrence~Zitnick, and Parikh]{vedantam2015cider}
R.~Vedantam, C.~Lawrence~Zitnick, and D.~Parikh.
\newblock Cider: Consensus-based image description evaluation.
\newblock In \emph{Proceedings of the IEEE conference on computer vision and pattern recognition}, pages 4566--4575, 2015.

\bibitem[Wang et~al.(2023{\natexlab{a}})Wang, Yang, Du, Fan, and Li]{wang2023clinicalgpt}
G.~Wang, G.~Yang, Z.~Du, L.~Fan, and X.~Li.
\newblock Clinicalgpt: Large language models finetuned with diverse medical data and comprehensive evaluation, 2023{\natexlab{a}}.

\bibitem[Wang et~al.(2023{\natexlab{b}})Wang, Zhong, Wang, Zhu, Mi, Wang, Shang, Jiang, and Liu]{wang2023data}
Z.~Wang, W.~Zhong, Y.~Wang, Q.~Zhu, F.~Mi, B.~Wang, L.~Shang, X.~Jiang, and Q.~Liu.
\newblock Data management for large language models: A survey, 2023{\natexlab{b}}.

\bibitem[Yang et~al.(2023)Yang, Xiao, Wang, Zhang, Bian, Yin, Lv, Pan, Wang, Yan, Yang, Deng, Wang, Liu, Ai, Dong, Zhao, Xu, Sun, Zhang, Liu, Ji, Xie, Dai, Fang, Su, Song, Liu, Ru, Ma, Wang, Liu, Lin, Nie, Guo, Sun, Zhang, Li, Li, Cheng, Chen, Zeng, Wang, Chen, Men, Yu, Pan, Shen, Wang, Li, Jiang, Gao, Zhang, Zhou, and Wu]{yang2023baichuan}
A.~Yang, B.~Xiao, B.~Wang, B.~Zhang, C.~Bian, C.~Yin, C.~Lv, D.~Pan, D.~Wang, D.~Yan, F.~Yang, F.~Deng, F.~Wang, F.~Liu, G.~Ai, G.~Dong, H.~Zhao, H.~Xu, H.~Sun, H.~Zhang, H.~Liu, J.~Ji, J.~Xie, J.~Dai, K.~Fang, L.~Su, L.~Song, L.~Liu, L.~Ru, L.~Ma, M.~Wang, M.~Liu, M.~Lin, N.~Nie, P.~Guo, R.~Sun, T.~Zhang, T.~Li, T.~Li, W.~Cheng, W.~Chen, X.~Zeng, X.~Wang, X.~Chen, X.~Men, X.~Yu, X.~Pan, Y.~Shen, Y.~Wang, Y.~Li, Y.~Jiang, Y.~Gao, Y.~Zhang, Z.~Zhou, and Z.~Wu.
\newblock Baichuan 2: Open large-scale language models, 2023.

\bibitem[Zeng et~al.(2023)Zeng, Liu, Du, Wang, Lai, Ding, Yang, Xu, Zheng, Xia, Tam, Ma, Xue, Zhai, Chen, Zhang, Dong, and Tang]{zeng2023glm130b}
A.~Zeng, X.~Liu, Z.~Du, Z.~Wang, H.~Lai, M.~Ding, Z.~Yang, Y.~Xu, W.~Zheng, X.~Xia, W.~L. Tam, Z.~Ma, Y.~Xue, J.~Zhai, W.~Chen, P.~Zhang, Y.~Dong, and J.~Tang.
\newblock Glm-130b: An open bilingual pre-trained model, 2023.

\bibitem[Zhong and Wang(2023)]{zhong2023chatgpt}
L.~Zhong and Z.~Wang.
\newblock Can chatgpt replace stackoverflow? a study on robustness and reliability of large language model code generation, 2023.

\end{thebibliography}

\appendix
% \section{Appendix}

\end{document}